\documentclass[10pt,twocolumn,letterpaper]{article}

\usepackage{cvpr}
\usepackage{times}
\usepackage{epsfig}
\usepackage{graphicx}
\usepackage{amsmath}
\usepackage{amssymb}
\usepackage{booktabs}
\usepackage{multirow}
\usepackage[activate={true,nocompatibility},final,tracking=true,kerning=true,spacing=true,factor=1100,stretch=10,shrink=10]{microtype}

\DeclareMathOperator*{\argmin}{arg\,min}
\cvprfinalcopy % *** Uncomment this line for the final submission

\def\cvprPaperID{3} % *** Enter the CVPR Paper ID here

\newcommand{\squeezeupSmall}{\vspace{-2mm}}
\newcommand{\squeezeup}{\vspace{-4mm}}

\ifcvprfinal\fi

\begin{document}

\title{ARCHANGEL:  Tamper-proofing Video Archives using \\Temporal Content Hashes on the Blockchain}

\author{T. Bui, D. Cooper, J. Collomosse\\
Centre for Vision Speech and Signal Processing\\
University of Surrey, UK.\\
\and
M. Bell, A. Green, J. Sheridan\\
The National Archives\\
Kew, London, UK.\\
\and
J. Higgins, A. Das, J. Keller, O. Thereaux\\
Open Data Institute (ODI)\\
London, UK.
\and
A. Brown\\
INDEX Business School,\\
University of Exeter, UK.\\}

\maketitle
\thispagestyle{empty}

%%%%%%%%% ABSTRACT
\begin{abstract}

We present ARCHANGEL; a novel distributed ledger based system for assuring the long-term integrity of digital video archives.  First, we describe a novel deep network architecture for computing compact temporal content hashes (TCHs) from  audio-visual streams with durations of minutes or hours.  Our TCHs are sensitive to accidental or malicious content modification (tampering) but invariant to the codec used to encode the video.  This is necessary due to the curatorial requirement for archives to format shift video over time to ensure future  accessibility.  Second, we describe how the TCHs (and the  models used to derive them) are secured via a proof-of-authority blockchain distributed across multiple independent archives. We report on the efficacy of ARCHANGEL within the context of a trial deployment in which the national government archives of the United Kingdom, Estonia and Norway participated. 

\end{abstract}

%%%%%%%%% BODY TEXT
\section{Introduction}

Archives are the lens through which future generations will perceive the events of today. Increasingly those events are captured in video form. Digital video is ephemeral, and produced in great volume, presenting new challenges to the trust and immutability of archives \cite{Johnson2013}.  Its intangibility leaves it open to modification (tampering) –- either due to direct attack, or due to  accidental corruption. Video formats rapidly become obsolete, motivating format shifting (transcoding) as part of the curatorial duty to keep content viewable over time.  This can result in accidental video truncation or frame corruption  due to bulk transcoding errors.

This paper proposes ARCHANGEL; a novel de-centralised system to guard against tampering of digital video within archives, using a permissioned blockchain maintained collaboratively by several independent archive institutions.  We propose a novel deep neural network (DNN) architecture trained to distil an audio-visual signature (content hash) from video, that is sensitive to tampering, but invariant to the format \ie the video codec used to compress the video.  Video signatures are computed and stored immutably within the ARCHANGEL blockchain at the time of the video's ingestion to the archive.  The video can be verified against its original signature at any time during curation, including at the point of public release, to ensure the integrity of content.  We are motivated by the need for national government archives to retain video records for years, or even decades.  For example, the National Archives of the United Kingdom receive supreme court proceedings in digital video form, which are held for five years prior to release into the legal precedent.  Despite the length of such videos it is critical that even small modifications are not introduced to the audio or visual streams. We therefore propose two technical contributions:

\noindent {\bf 1) Temporal Content Hashing.}  Cryptographic hashes (\eg SHA-256 \cite{sha}) operate at the bit level, and whilst effective at detecting video tampering are not invariant to transcoding.  This motivates {\em content-aware} hashing of the audio-visual stream within the video.  We propose a novel DNN based temporal content hash (TCH) that is trained to ignore transcoding artifacts, but is capable of detecting tampers of a few seconds duration within typical video clip lengths in the order of minutes or hours.  The hash is computed over both the audio and the visual components of the video stream, using a hybrid CNN-LSTM network (c.f. subsec.~\ref{sub:arch}) that is trained to predict the frame sequence and thus learns a spatio-temporal  representation of clip content.  

\noindent {\bf 2) Cross-archive Blockchain for Video Integrity}  We propose the storage of TCHs within a permissioned proof-of-authority blockchain maintained across multiple independent archives participating in ARCHANGEL, enabling mutual assurance of video integrity within their archives. Fine-grain temporal sensitivity of the TCH is challenging given the requirement of hash compactness for scalable storage on the blockchain.  We therefore take a hybrid approach. The codec-invariant TCHs computed by the model are stored compactly on-chain, using product quantization \cite{Jegou2010}.  The model itself is stored off-chain alongside the source video (\ie within the archive), and hashed on-chain via SHA-256 to mitigate attack on the TCH computation.  

The use of Blockchain in ARCHANGEL is motivated by changing basis of public trust in society. Historically, an archives' word was authoritative, but we are now in an age where people are increasingly questioning institutions and their legitimacy.  
Whilst one could create centralised database of video signatures, our work enables a shift from an institutional underscoring of trust, to a technological underscoring of trust.  The cryptographic assurances of Blockchain evidence trusted process, whereas a centralised authority model reduces to  institutional trust.

We demonstrate the value of ARCHANGEL through a trial international deployment across the national government archives of the United Kingdom, Norway and Estonia.  Collaboration between the majority of partnering archives (`50\% attack' \cite{Narayanan2016}) would be needed to rewrite the blockchain and so undermine its integrity, which is unlikely between archives of independent sovereign nations.  This justifies our use of Blockchain, which to the best of our knowledge, has not been previously combined with temporal content-hashing to ensure video archive integrity.

%%%%%%%%%%%%%%%%%%%%%%%%%%%%%%%%%%%%%%%%%%%%%%%%%%%%%%%%%
\section{Related work}

Distributed Ledger Technology (DLT) has existed for over a decade - notably as blockchain, the technology that underpins Bitcoin \cite{bitcoin}.  The unique capability of DLT is to guarantee the integrity and provenance of data when distributed across many parties, without requiring those parties to trust one another nor a central authority \cite{Narayanan2016}; \eg Bitcoin securely tracks currency ownership without relying upon a  bank. Yet, the guarantees offered by DLT for distributed, tamper-proof data have potential beyond the financial domain \cite{Walport2015,Holmes2018}.   We build upon recent work suggesting the potential of DLT for digital record-keeping \cite{archangel,Lemieux2016}, notably an earlier incarnation of ARCHANGEL described in Collomosse \etal \cite{archangel} which utilised a proof-of-work (PoW) blockchain to store SHA-256 hashes of binary zip files containing academic research data.  Also related, is the prior work of Gipp \etal \cite{Gipp2016} where SHA-256 hashes specifically of video are stored in a PoW (Bitcoin) blockchain for evidencing car collision incidents.  Our framework differs, due to the insufficiency of such binary hashes \cite{sha} to verify the immutability of video content in the presence of format shifting, which in turn demands novel content-aware video hashing, and a hybrid on- and off-chain storage strategy for safeguarding those hashes (and DNN models generating them).

Video content hashing is a long-studied problem, \eg tackling the related problem of near-duplicate detection for copyright violation \cite{Googlepatent}.  Classical approaches to visual hashing explored heuristics to extract video feature representations including spectral descriptors \cite{Khelifi17}, and robust temporal matching of visual structure \cite{Cao2012,Ye2013,Song2013} or sparse gradient-domain features  \cite{Sivic2003,Sivic2007} to learn stable representations. The advent of deep learning delivered highly discriminative hashing algorithms \eg derived from deep auto-encoders \cite{Song2018,zhang2016play}, convolutional neural networks (CNN) \cite{Zu2015,Liong17,kim2018self} or recurrent neural networks \cite{Gu2016,Wu2017}.  These technique train a global model for content hashing over a representative video corpus, and focus on hashing short clips with lengths of a few minutes at most using visual cues only.  Audio-visual video fingerprinting via multi-modal analysis is sparsely researched with prior work focusing upon augmenting visual matching with an independent audio fingerprinting step \cite{Googlepatent}, predominantly via wavelet analysis \cite{Baluja2008,Senoussaoui2014}.  Our archival context requires hashing of diverse video content; training a single representative DNN to hash such video is unlikely practical nor future-proof.  Rather our work takes an unsupervised approach, fitting a triplet CNN-LSTM network model to a single clip to minimise frame reconstruction and prediction error.  This has the advantage of being able to detect short duration tampering within long duration video sequences (but at the overhead of storing a model per hashed video).  Since videos in our archival context may contain limited visual variation (\eg a committee or court hearing) we hash both video and audio modalities.  Uniquely, we expose the model to variations in video compression during training to avoid detecting transcoding artifacts as false positives.

\begin{figure}
    \centering
    \includegraphics[width=\linewidth]{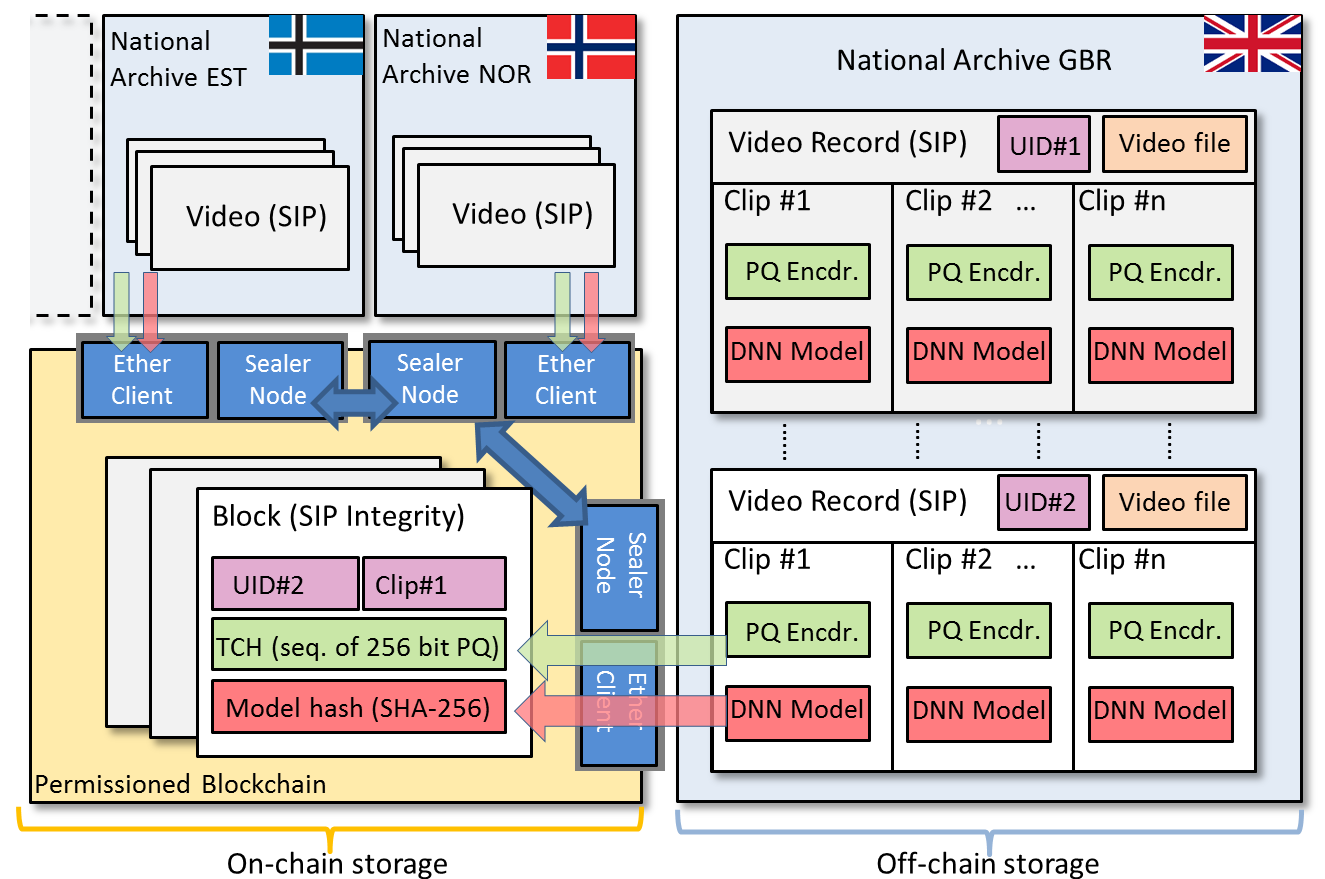}
    \caption{ARCHANGEL Architecture.  Multiple archives maintain a PoA blockchain (yellow) storing hash data that can verify the integrity of video records (video files and their metadata; sub-sec.~\ref{sec:blockchain}) held off-chain, within the archives (blue). Unique identifiers (UID) link on- and off-chain data. Two kinds of hash data are held on-chain: 1) A temporal content hash (TCH) protecting the audio-visual content from tampering, computed by passing sub-sequences of the video through a deep neural network (DNN) model to yield a sequences of short binary (PQ \cite{Jegou2010}) codes (green); 2) A binary hash (SHA-256) of the DNNs and PQ encoders that guards against tampering with the TCH computation (red).}
    \label{fig:data-arch}
    \squeezeup
    \squeezeupSmall 
\end{figure}

The emergence of high-realism video manipulation using deep learning ('deep fakes' \cite{Kautz2017}) has renewed interest in determining the provenance of digital video.  Most existing approaches focus on the task of detecting visual manipulation\cite{deepfakedet}, rather than immutable storage of video hashes over longitudinal time periods, as here.  Our work does not tackle the task of  detecting video manipulation {\em ab initio} (we trust video at the point of ingestion).  Rather, our hybrid AI/DLT approach can detect subsequent tampering with a video, offering a promising tool to combat this emerging threat via proof of content provenance.

\section{Methodology}
\begin{figure*}
    \centering
    \includegraphics[width=0.8\linewidth,height=6cm]{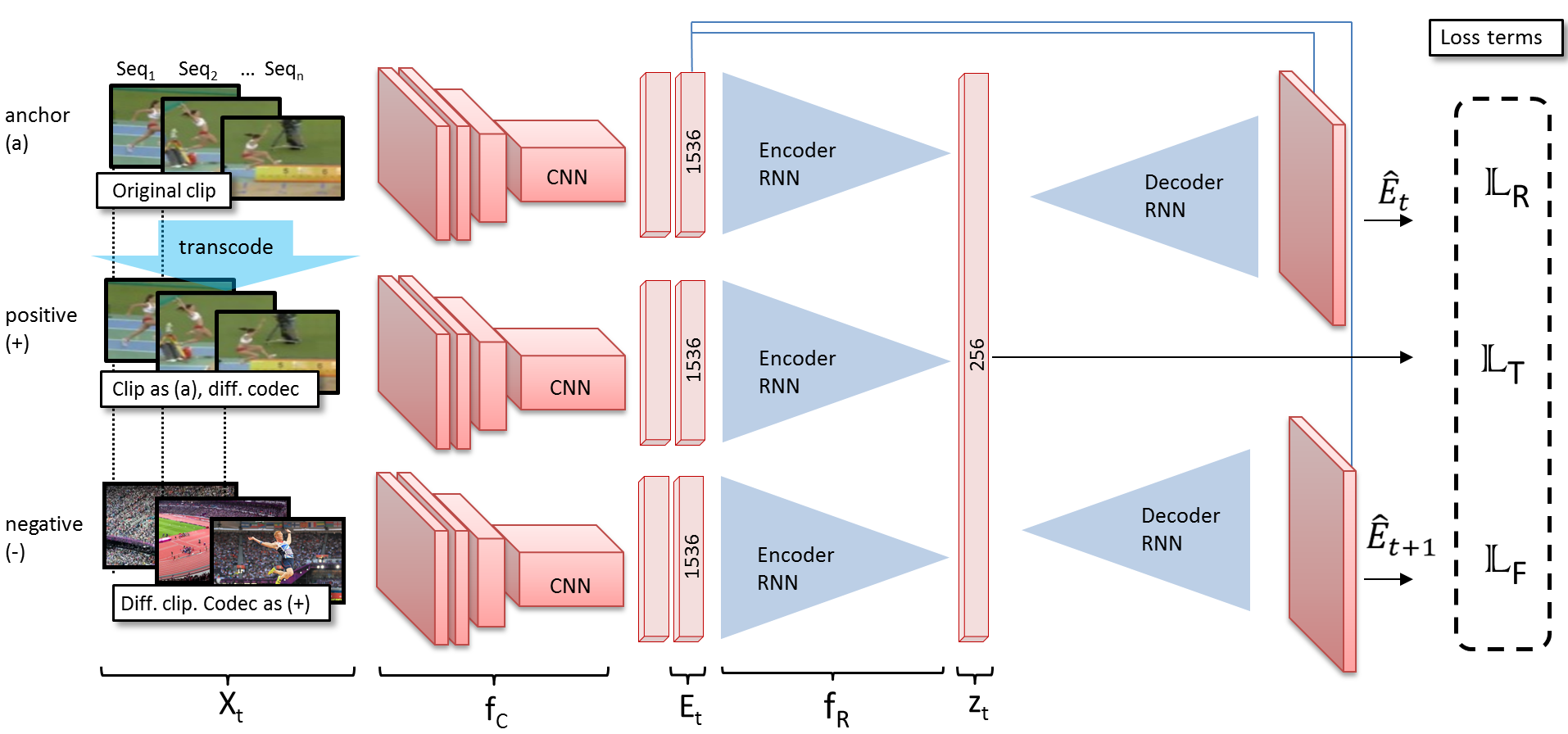}
    \caption{Our triplet auto-encoder for video tamper detection. This schematic is for visual cues, using InceptionResNetV2 backbone to encode video frames (audio network uses MFCC features).  Once extracted, features are passed through the RNN encoder-decoder network which learns invariance to codec but to discriminate between differences in content.  Bottleneck $z_t$ serves as the content hash and is compressed via PQ post-process (not shown).  Video clips are fragments into 30 second sequences each yielding a content hash (PQ Code). The temporal content hash (TCH) stored on-chain is the sequence concatenation of the PQ code strings for both audio/video modalities.  }
    \label{fig:arch}
\end{figure*}

We propose a proof-of-authority (PoA, permissioned) blockchain as a mechanism for assuring provenance of digital video stored off-chain, within the public National Archives of several nation states. The blockchain is implemented via Ethereum \cite{eth} maintained by those independently participating archives, each of which runs a sealer node in order to provide consensus by majority on data stored on-chain.  Fig.~\ref{fig:data-arch} illustrates the architecture of the system.  We describe the detailed operation of the blockchain in subsec.~\ref{sec:blockchain}.  The integrity of video records within the archives is assured by content-aware hashing of each video file using a deep neural network (DNN).  The DNN is trained to ignore modification of the video due to change in video codec (format) but to be sensitive to tampering. The DNN architecture and training are described in subsecs.~\ref{sub:arch}-\ref{sub:loss} and the hashing and detection process in subsec.~\ref{sub:cluster}-\ref{sub:tamp}.

\subsection{Video Tamper Detection}
\label{sec:ml}

We cast video tamper detection as an unsupervised representation learning problem.  A recurrent neural network (RNN) auto-encoder is used to learn a predictive spatio-temporal model of content from a single video.  A feed-forward pass of a video sequence generates a sequence of compact binary hash codes.  We refer to this sequence as a {\em temporal content hash} (TCH).  The RNN model is trained such that even minor spatial (frame) corruption or temporal discontinuities (splices, truncations) in content will generate different TCH when fed forward through the model, yet transcoding content will generate near-identical TCH.  

The advantage of learning (over-fitting) a video-specific model to content (rather than adopting a single RNN model for encoding all video) is that a performant model can be learned for very long videos without introducing biases or assumptions on what constitutes representative content for an archive; such assumptions are unlikely to hold true for general content.  The disadvantage is that a model must be stored with the archive, as meta-data alongside the video.  This is necessary to ensure the integrity of the model so  preventing attacks on the TCH verification process.
\squeezeupSmall
\subsubsection{Network Architecture}
\label{sub:arch}

We adopt a novel triplet encoder-decoder network structure resembling an RNN autoencoder (AE).  Fig.~\ref{fig:arch} illustrates the architecture for encoding visual content.

Given a continuous video clip $\mathcal{X}$ of arbitrary length we sample keyframes at 10 frames per second (fps). Keyframes are aggregated into $B=\frac{|\mathcal{X}|}{N}$ blocks where $N=300$ frames ($\sim$ 30 seconds, padded with empty frames if necessary in the final block), yielding a series of sub-sequences $\mathcal{X}=\{X_1,X_2,...,X_B\} \in \mathbb{R}^{N\times H\times W\times C}$ where H, W and C are the height, width and number of channels for each image frame. The encoder branch is a hybrid CNN-LSTM (Long-Short Term Memory) encoder function in which an InceptionResNetV2 CNN backbone encodes each block $X_t$ to a sequence of semantic features that summarize spatial content.  $f_C : \mathbb{R}^{H\times W \times C} \to \mathbb{R}^P$, which encodes $\mathcal{X}$ into $E=\{E_t \in \mathbb{R}^{N \times S}~|~t=0,1,...,B \}$; for InceptionResNetV2 \cite{szegedy2017inception} $S=1536$ from the final fc layer.

The LSTM learns a deep representation encoding the  temporal relationship between frame features by encoding E with a RNN, $f_R : \mathbb{R}^{N\times S} \to \mathbb{R}^D$, resulting in an embedding $Z=\{z_t \in \mathbb{R}^D~|~t=0,1,...,B\}$. The learning of Z is constrained by a set of objective functions described in subsec.~\ref{sub:loss} ensuring the representatives, continuity and transcoding invariance of the learned features. 

Audio features are similarly encoded via triplet LSTM but omitting the CNN encoder function $f_C$ and substituting MFCC co-efficients from the audio track of each block $X_t$ as features directly. We later demonstrate that encoding audio is critical in the cases where video contains mostly static scenes e.g. a video account of a meeting or court proceedings.  The AE bottleneck for audio features is 128-D; we denote this embedding $z'_t$.

\subsubsection{Training Methodology}
\label{sub:loss}

Given video clip $\mathcal{X}$ we train $f_R(f_C(.))$ independently yielding a pair of models $(M,M')$ for the video and audio modalities, minimizing a training loss comprising three equally-weighted terms \eg for all blocks $X_t \in \mathcal{X}$:
\begin{eqnarray}
M_{(\mathcal{X})}=\argmin_\theta \mathcal{L}_R(X_t;\theta)+\mathcal{L}_F(X_t;\theta)+\mathcal{L}_T(X_t;\theta).
%M(\mathcal{X})=\sum_{\forall X_t \in \mathcal{X}} \argmin_\theta \mathcal{L}_R(X_t;\theta)+\mathcal{L}_F(X_t;\theta)+\mathcal{L}_T(X_t;\theta).
\end{eqnarray}

\noindent \textbf{Reconstruction loss}, $\mathcal{L}_R$, aims to reconstruct input sequence $\hat{E}_t$ from $z_t$ that approximates $E_t$. This loss is measured using Mean Square Error (MSE), effectively makes the network an auto-encoder as in~\cite{zhang2016play}.
\begin{eqnarray}
\mathcal{L}_R = \frac{1}{2}|E_t - \hat{E}_t|^2_2.
\end{eqnarray}
\noindent \textbf{Prediction loss}, $\mathcal{L}_F$, aims to predict the next sequence $\hat{E}_{t+1}$ from $z_t$ that approximates $E_{t+1}$ via MSE. While the reconstruction loss ensures the integrity within sequences, the prediction loss provides inter-sequence links which is important for encoding very long videos. For the final sequence $t = B$, this loss term is simply turned off. This work is similar to~\cite{kim2018self} however they used 3D CNN for spatio-temporal encoding instead of LSTM in our work. 
\begin{eqnarray}
\mathcal{L}_F = \frac{1}{2}|E_{t+1} - \hat{E}_{t+1}|^2_2.
\end{eqnarray}

\noindent \textbf{Triplet loss}, $\mathcal{L}_T$, brings together similar sequences $(z_a,z_p)$ while pushing dissimilar sequences $(z_a,z_n)$ such that their difference is below a certain margin:
\begin{equation}
    \mathcal{L}_T= \frac{1}{\left | T \right |}\sum_{(z_a, z_p,z_n) \in T}{\{d(z_a,z_n) - d(z_a,z_p) + m\}_+}
    \label{eq:triplet}
\end{equation}
where $d(.)$ is the cosine similarity metric and $\{.\}_+$ is the hinge loss function. Here we desire the embedding Z invariant to changes in video formats (\eg codec, container, compression quality) hence the positive sample, $X_t^p$, is set to be the same video sequence as the original sequence $X_t$ but transcoded differently $z_p = f_R(f_C(X_t^p))$. To generate $X_t^p$ we augment $\mathcal{X}$ using various transcoding combinations (details are described in subsec.~\ref{sub:data}), resulting in a much larger training set $\mathcal{X}^P$ that also helps to reduce overfitting. To make the learning harder, the negative sample is selected among other sequences having the same encoding format as the anchor $z_n = f_R(f_C(X^n_{t*})),~X^n_{t*} \in \mathcal{X}\backslash X_t$. The margin m is fixed $m=0.2$ in our experiments.  The network is trained end-to-end via the Stochastic Gradient Descent optimizer (SGD with momentum, decay learning rate and early stopping following~\cite{wilson2017marginal}) with InceptionResNetV2 initially pre-trained over the ImageNet corpus~\cite{szegedy2017inception}.

\subsubsection{Clip detection and clustering}
\label{sub:cluster}
 Archival videos are typically formed from multiple visually distinct parts (\eg titles, attributions, shots captured from different stationary cameras). It is undesirable to train an RNN to model continuity within a single video where potentially several such scene changes are present. Without loss of generality we therefore propose to split the video into multiple clips based on changes in frame continuity using MPEG-4 standard; a Sum of Absolute Difference (SAD) \cite{richardson2004h} operator. Any other scene cut detector could be substituted.

This results in the input video being split into several, sometimes hundreds, of different clips. Such clips are independently processed as $\mathcal{X}$, per the above.  Since training a model for each clip is undesirable, we wish to group similar clips together and train a model for each group. To do so we sample $\sim 100$ key frames from each scene  extract semantic features from these frames (using $f_C(.)$) and cluster them into $K$ groups. Clip $\mathcal{X}$ belongs to cluster $k$ if the majority of its frames are assigned to $k$. We use Meanshift for clustering \cite{comaniciu2002mean}; $K$ is decided automatically (typically 2-3) per video.

\subsubsection{Video hashing and tampering detection}
\label{sub:tamp}
Given a video $\mathcal{X}$, the representation $z_t$ of sub-sequence $X_t$ where $t=[0,B]$ can be obtained by passing $X_t$ through its respective spatio-temporal model $f_R(f_C(.))$. Additionally, for each scene we define a threshold value set to tolerate variation due to video transcoding:
\begin{equation}
    \epsilon_t = \max_{z_t^p \in Z_t^p}{\left|\left| z_t - z_t^p \right| \right|_2}
    \label{eq:thres}
\end{equation}
where $Z_t^p$ is the collection of bottleneck representations of the positive clips (\ie transcoded) $X^p_t$ used to form the training triplets. $\epsilon_t$ should tolerate any video transcoding in future with assumption that future codecs are likely to encoder with higher fidelity than  existing  methods. 

The collective representations of all clips that forms $\mathcal{X}$, along with other meta info such as threshold $\epsilon_t$ and clip IDs are stored in blockchain as the content hash of video $\mathcal{X}$. As $\mathcal{X}$ can have an arbitrary length, its content hash could exceed the current blockchain transaction limit. To address this problem, we further compress Z into 256-bit binary using Product Quantisation (PQ)~\cite{Jegou2010}. PQ requires the training of a further model (referred to here as `PQ Encoder') capable of compressing $Z \in \mathbb{R}^{256}$ to $\zeta \in \mathbb{B}^{256}$ and estimating the $L_2$ norm between pairs of such feature representations.

At verification time, we determine the integrity of a query video by feeding forward each constituent clip $X^*$ through the model pair $(M,M')$ trained for $\mathcal{X}$, to obtain its video and audio TCHs. These hashes of $\mathcal{X}^*$ are compared against the corresponding hashes of $\mathcal{X}$ stored within the blockchain. $\mathcal{X}^*$ is considered tampered if there exists a pair of hashes whose distance greater than the corresponding threshold ($\epsilon_t$). This method is advantageous as it provides a localised indication of which clip within the video experienced tampering. 

\begin{figure}[t!]
    \centering
    \includegraphics[width=\linewidth,height=5cm]{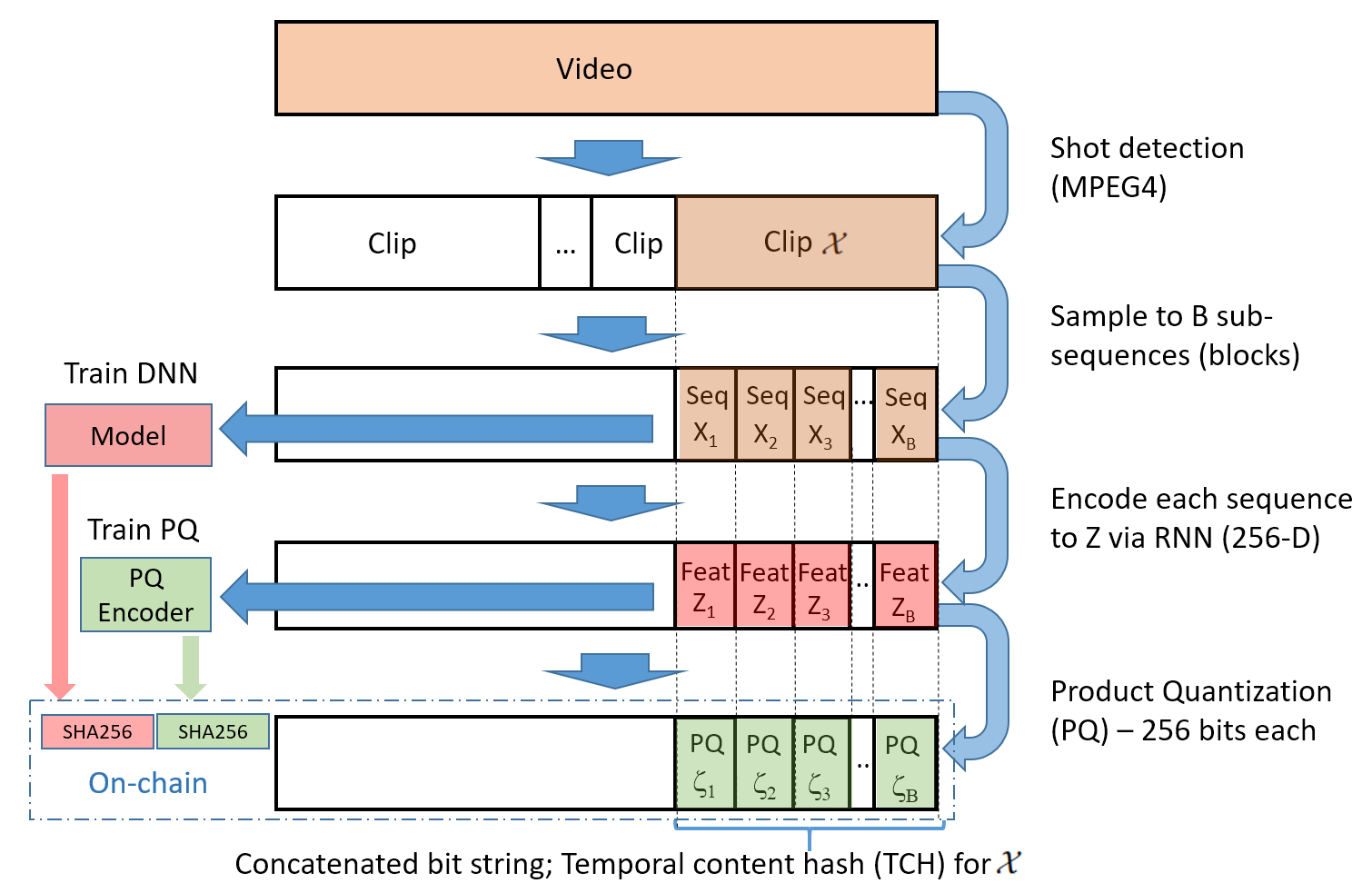}
    \caption{Schematic of TCH Computation.  Videos are split into clips which are each independently hashed to a TCH.  The TCH is the concatenation of PQ codes that are derived from RNN bottleneck features ($z_t$).  Each code hashes a block of $N$ frames. Encoding pipeline shown for visual cues only; audio is similarly encoded.}
    \label{fig:encoder}
            \squeezeupSmall

\end{figure}

\subsection{Ensuring Hash Integrity}
\label{sec:blockchain}

Hashing each clip within a video process yields its TCH; a sequence of binary hashes derived from the visual content: $\{X_1,X_2,...,X_B\} \mapsto \zeta=\{\zeta_1,\zeta_2,...,\zeta_B\}$ and similarly a sequence of audio hashes $\zeta'=\{\zeta'_1,\zeta'_2,...,\zeta'_B\}$.  These hashes are stored immutably on the blockchain alongside a universally unique identifier (UID) for the clip.  To safeguard the integrity of the verification process, it is necessary to also store a hash of the model pair $(M,M')$ and threshold value $\epsilon_t$ also on the blockchain (Fig.~\ref{fig:encoder}).  Without loss of generality, we compute the model hashes via SHA-256.  Use of such a bit-wise hash is valid for the model (but not the video) since the model (which is a set of filter weights) remains constant but the video may be transcoded, during its lifetime within the archive.  The proposed system requires that the model pair be stored off-chain due to storage size limitations (32Kb per transaction in our Ethereum implementation), and the potential for reversibility attack on the model itself which might otherwise apply generative methods \cite{carlini2018secret} to enable partial disclosure of timed-release or non-public archival records.  Such reversals are not feasible for the 256 bit PQ hashes stored on the chain, particularly given the off-chain storage of the PQ and DNN models.

Practically, our system is implemented on a compute infrastructure run locally at each participating archive.  The system has been designed with archival practices in mind, utilising data packages conforming to the Open Archival Information System (OAIS) \cite{oais}, and as such the archivist user must provide metadata for an archive data package (Submission Information Package (SIP)) that the video files are part of.  A web-app run locally inspects each SIP as it is submitted where it is verified to be a suitable file type, and stored within the locally held archive (database) for processing by a background daemon.  The daemon trains model pair $(M,M')$ and the PQ encoder on a secure high performance compute cluster (HPC) with nodes comprising four NVidia 1080ti GPU cards.  The time taken to train the model is typically several hours, although subsequent TCH computation takes less than one minute.  These timescales are acceptable given the one-off nature of the hashing process for a video record likely to remain in the archive for years or decades.

This data is then submitted as a record to the blockchain by way of a smart contract transaction which manages the storage and access of data. In order to submit data the user must be given access via the smart contract, which the application interfaces (APIs) then use to manage access to the submission functionality. Once the transaction is processed, the UIDs for the SIP and the video files can be used to access data stored on the chain for subsequent verification.  For reasons of sustainability and scalability of the system, we opted on a proof-of-authority (PoA) consensus model. Under such a model, there is no need for computationally expensive proof of work mining.  Rather, blocks are sealed at regular intervals by majority consensus of the clique of nodes pre-authorised to.  In our trial deployment over a small number of archives, access keys were pre-granted and distributed by us.  In a larger deployment (or to scale this trial deployment) the PoA model could be used to grant access via majority vote to new archives or memory institutions wishing to join the infrastructure.

%%%%%%%%%%%%%%%%%%%%%%%%%%%%%%%%%%%%%%%%%%%%%%%%%%%%%%%%%
\section{Experiments and Discussion}
\label{sec:eval}

We evaluate the performance of our system deployed for trial across the National Archives of three nations: the United Kingdom, Norway and Estonia.  We first evaluate the accuracy of video tamper detection algorithm and then report on the performance characteristics of that deployment.

\subsection{Datasets and Experimental Setup}
\label{sub:data}
\begin{figure}
    \centering
    \includegraphics[width=\linewidth,height=5cm]{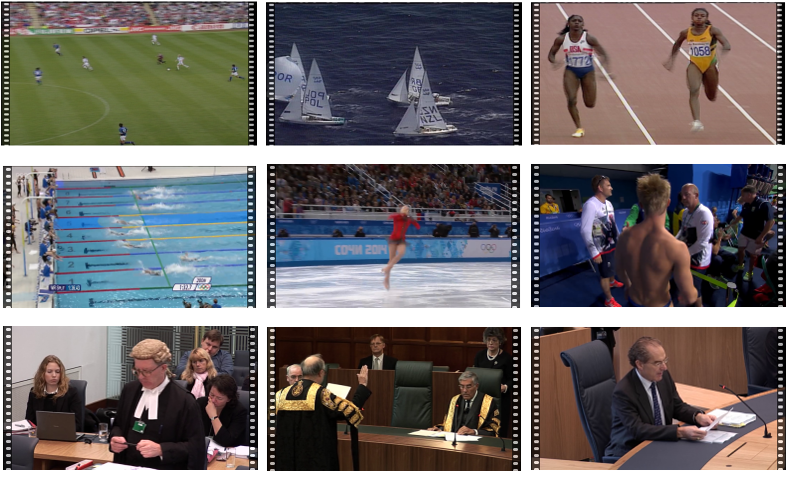}
    \caption{Representative thumbnails of the long-duration video datasets used in our evaluation.  Top: ASSAVID, uneditted `rushes' footage from the BBC archives of the 1992 Olympics; Middle:  OLYMPICS, editted footage from Youtube of the 2012-2016 Olympics; Bottom: TNA, Court hearings --- mostly static visuals.}
    \label{fig:data}
            \squeezeupSmall

\end{figure}

Most video datasets in the literature are of short clips \ie few seconds to several minutes of relatively narrow domain. However in archiving practice, a video record of an event can be of arbitrary length; typically tens of minutes or even hours. We evaluate with 3 diverse datasets more representative of an archival context:

\noindent \textbf{ASSAVID}~\cite{messer2002automatic} contains 21 television broadcast videos taken from Barcelona Olympics of 1992. This dataset covers various sports with commentary and music in the audio track.  Video length ranges from 12 seconds to 40 minutes, encoded using the older MPEG-2 video and MP2 audio codecs.

\noindent \textbf{OLYMPICS} contains 5 fast-moving Youtube videos taken from Olympics 2012-2016 covering final rounds of competitions in 5 different sports including running, swimming, skating and springboard. The dataset has modern MPEG-4 encoding. The average length of videos is 10 minutes.

\noindent \textbf{TNA} contains 7 court hearing videos released by the United Kingdom National Archives. The video length ranges from 4 minutes to 2 hours, averagely 55 minutes per video. This is a challenging dataset as the videos contains highly static scenes of courtroom but salient audio track. Examples of these datasets are shown in Fig.~\ref{fig:data}.

For each video within the three datasets above, we create three test sets: a `{\em control}' set, a `{\em temporal tamper}' set and a `{\em spatial tamper}' set. The `{\em control}' testset contains 10 duplicates of each video, each being a transcoded version of the original one. Specifically, we use the modern video codec H.265 with random frame rate (26/48/60 fps), 12 levels of compression (controlled by the parameter {\em crf} under ffmpeg) and container (.mp4, .mkv) to transcode the videos. Audio is also transcoded to a new codec (libmp3lame, aac) with random sampling rate and compression quality. In the worst scenario the transcoded video could be 5 times smaller in file size than the original video. The `{\em temporal tamper}' test set contains 100 videos generated from each original video by randomly removing a chunk of 1-10 seconds while keeping the same encoding formats. Similarly, the `{\em spatial tamper}' test set contains 100 videos where frames/audio within random sections of the video (1-10 seconds in length) are replaced by white noise.  

\noindent \textbf{Experimental settings}  
% CNN/RNN config, embedding dim, PQ, HxW resize
% augmentation
We experimented with 3 CNN architectures for $f_C$ -- VGG~\cite{simonyan2014very}, ResNet~\cite{he2016deep} and InceptionResnetV2~\cite{szegedy2017inception} using the embedding provided by the final fully connected/pooling layer as the output to $E$. To minimise model size we freeze $f_C$ and only train $f_R$. The RNN encoder is a LSTM network with 1024 cells while the reconstruction and prediction decoders have 2048 cells each. The dimension of embedding vector $z$ is empirically set 256-D for video and 128-D for audio prior to PQ. Input frames are scaled down to fit the corresponding CNN input size \ie 299x299 for InceptionResNetV2 and 224x224 for VGG/ResNet. Input audio is converted (if necessary) to a mono stream for our experiments. Once the model has been trained we save the encoder part of the network off-chain; the off-chain storage required to verify a clip is approximately 100Mb total for audio and video models. %Videos in our test set typically yield 2-4 clip clusters regardless of its quality or video length.

\noindent \textbf{Data augmentation} in the form of video transcoding, is used to form the triplets used to train our network in order to develop codec invariance. For each video we created 50 different transcoded versions using current or older codecs (\eg H.264, H.263p, VP8) at a variety of frame rates from 10fps to 25fps and with a variety of compression/quality parameters spanning typical high and low values for each codec.  The transcoder used was the open-source {\em ffmpeg} library with `compression factor' an integer number reflecting presets for the codecs provided with that library. Note that these codec sets differ from those natively used in the test set.  These transcoded examples form the positive exemplars for the triplet network (subsec.~\ref{sub:loss}). 

\subsection{CNN architecture and loss ablation study}

\begin{figure}[t!]
    \centering
    \begin{tabular}{c}
        \includegraphics[width=\linewidth]{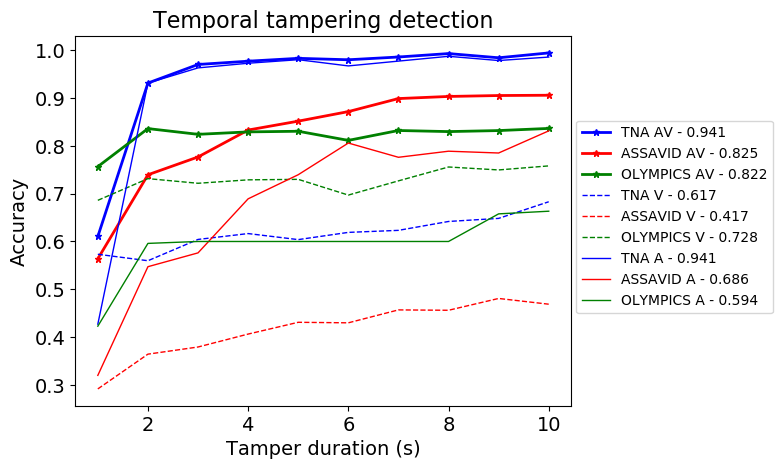}  \\
        \includegraphics[width=\linewidth]{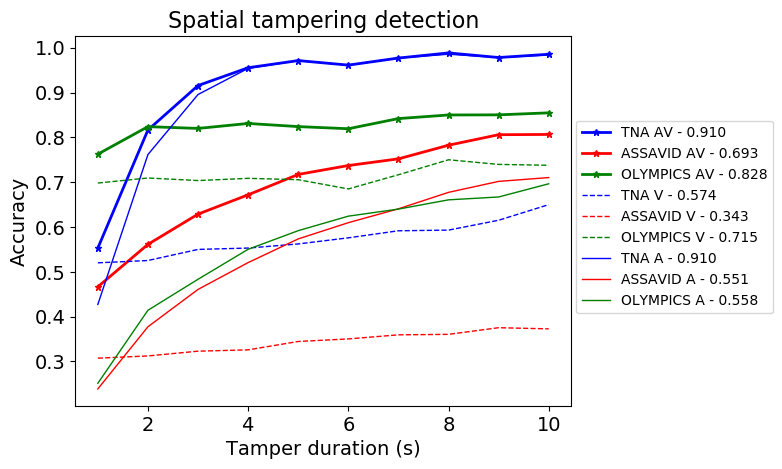}
    \end{tabular}
    \caption{Plotting accuracy of tamper detection vs. video length for the temporal (top) and spatial (bottom) tampering cases for each of the three sets.  Ablations of the system using audio and visual modalities only are indicated via suffix (-A, -V).}
    \label{fig:tamlen}
\end{figure} 

\begin{table}[t!]
    \centering
    \small
    \begin{tabular}{|c|c|c|c|}
    \hline
        Architecture & Temporal & Spatial & Control \\ \hline
        VGG-16~\cite{simonyan2014very} & 0.876 & 0.753 & 0.909 \\ \hline
        ResNet50~\cite{he2016deep} & 0.856 & 0.747 & 0.913 \\ \hline
        InceptionResNetV2~\cite{szegedy2017inception} & {\bf 0.898} & {\bf 0.809} & {\bf 0.955} \\ \hline
    \end{tabular}
    \caption{Evaluating visual tamper detection performance for various backbones in our CNN-LSTM architecture.  Values expressed are accuracy rates for {\em spatial} and {\em temporal} tamper detection, and true negative rates for {\em control} averaged across all three test sets. }
    \label{tab:cnn}
            \squeezeupSmall

\end{table}
We experimented with different CNN architectures for the encoder backbone, and with ablations of our proposed loss function on a subset of 12 videos sampled from all datasets above. Table.~\ref{tab:cnn} shows the tamper detection accuracy on the {\em temporal} and {\em spatial} tamper sets, as well as the true negative rate for the control sets.  When varying the CNN backbone all other parts of the network are kept the same. While there is no clear margin between VGG and ResNet, the InceptionResNetV2 architecture shows superior performance on all test sets.  

We perform ablations on the three terms comprising total loss (eq.~\ref{eq:triplet}) while keeping the network, data and training parameters unchanged. The contribution of each loss to tamper-detection capability is reflected in Table.~\ref{tab:loss_contr}. The reconstruction loss alone (\ie AE model similar to~\cite{zhang2016play}) has best score on the control set since its sole objective is to reconstruct the input sequences. The lack of learning coherence between adjacent sequences causes it to under-perform detection on both of the tamper sets. In contrast, the prediction loss (similar to~\cite{kim2018self}) helps to learn an embedding more sensitive to tampering and the triplet loss balances the performance on all three test sets.

\begin{table}
    \centering
    \small
    \begin{tabular}{|c||c|c|c||c|c|c|}
    \hline
        Method & $\mathcal{L}_R$ & $\mathcal{L}_F$ & $\mathcal{L}_T$ & Temporal  & Spatial & Control\\ \hline
        \cite{zhang2016play} & $\bullet$ &   &   & 0.884 & 0.749 & {\bf 0.968} \\  \hline
        \cite{kim2018self} & $\bullet$ & $\bullet$ &   & 0.896 & 0.776 & 0.945 \\ \hline
        Ours & $\bullet$ & $\bullet$ & $\bullet$ & {\bf 0.898} & {\bf 0.809} & 0.955 \\ \hline
    \end{tabular}
    \caption{Ablation experiment exploring term influence on total loss (eq.~\ref{eq:triplet}): $\mathcal{L}_R$ (reconstruction), $\mathcal{L}_F$ (prediction) and $\mathcal{L}_T$ (triplet). Solid dot indicates incorporation of loss term.  Values expressed are accuracy rates for {\em spatial} and {\em temporal} tamper detection, and true negative rates for {\em control} averaged across all three test sets.}
    \label{tab:loss_contr}
            \squeezeupSmall

\end{table}

\subsection{Evaluation of tampering detection}

We evaluated the performance of tamper detection for all three test sets for a variety of tamper lengths.  Fig.~\ref{fig:tamlen} shows the trend of tamper length to detection accuracy for the propose system and also on an ablated version of the system reliant solely on audio and solely on video modalities.  on the performance of the audio and video models. Both {\em temporal} and {\em spatial} tamper sets have a similar trend:  models are consistently effective at detecting tampered  segments of around 3 seconds or more.  Also this is excellent sensitivity given input videos lasting tens of minutes or hours (see subsec.\ref{sub:data}) the system struggles to detect tampering of less than 2 seconds. This may be due to the transcoding of videos during the training data augmentation causing frame loss for certain combinations of codec, frame rate and container, or the inherent challenge of achieving further sensitivity given such video lengths. Fig.~\ref{fig:tamlen} illustrates the importance of encoding both video and audio together for all test sets. Audio contributes significantly to the tampering detection performance on the TNA dataset (where visual content of the courtroom hearing is highly static but audio is clear and salient) and ASSAVID (where low visual quality and older codec make the reading of frames difficult). On the other hand, the OLYMPICS dataset has excellent visual quality and dynamic motion which benefit the video models. In contrast, all audio streams from the OLYMPICS dataset have either music (sub-optimal for MFCC) or loud background noise that downplays its audio models. As with audio-visual combination the tampering detection accuracy is significantly boosted on all datasets.

\begin{figure}
    \centering
    \includegraphics[width=0.9\linewidth,height=5cm]{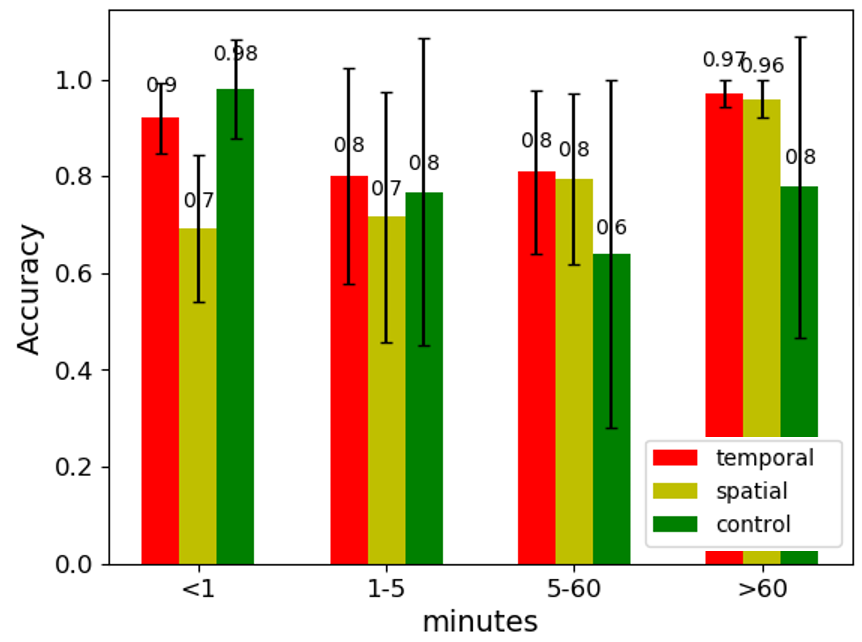}
    \caption{Overall accuracy vs. video length averages across all three test sets. Values expressed are accuracy rates for {\em spatial} and {\em temporal} tamper detection, and true negative rates for {\em control} averaged across all test sets partitioned by length.}
    \label{fig:vidlen}
            \squeezeupSmall
\end{figure}

Fig.~\ref{fig:vidlen} aggregates performance statistics across all datasets grouping videos by length, which is shown not to affect the detection performance. Approximately 10\% of the videos have length greater than 1 hour, yet over 95\% of the tampers present in their test sets are detected correctly. Additionally, Fig.~\ref{fig:vidlen} also shows that {\em spatial} tamper is harder to detect than {\em temporal}.  This may be due to the spatial features being embedded by a pre-trained model (InceptionResNetV2) providing some resilience to noise.

\begin{figure}
    \centering
    \includegraphics[width=\linewidth]{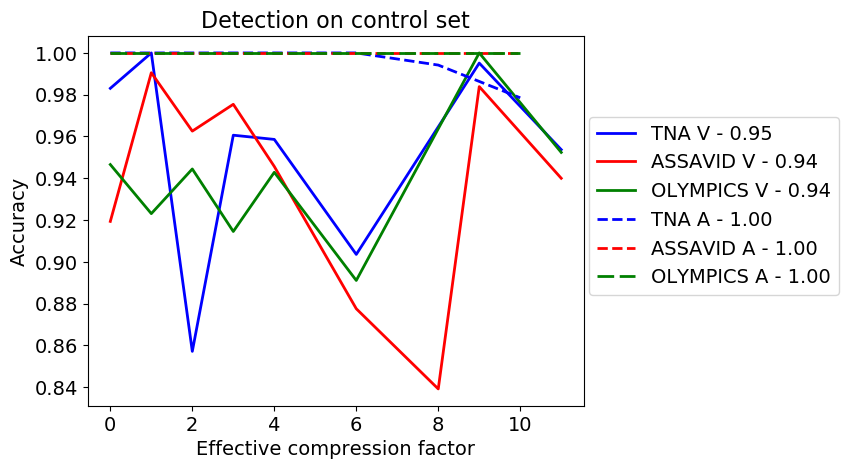}
    \caption{Plotting true negative rate over the control set (containing transcoded, untampered videos). Videos have been downgraded in their visual encoding (-V) and audio encoding (-A).  For visualisation purpose we scale the video and audio compression factors input to {\em ffmpeg} to the same range. There is no observable trend between compression quality and false tamper detection rate.}
    \label{fig:compress}
            \squeezeupSmall

\end{figure}

We also investigated the effect of the compression factor in the control set in Fig.~\ref{fig:compress}; \ie the ability of the system to reflect true negatives (avoid false detections). Data augmentation during training (subsec.~\ref{sub:loss} and \ref{sub:data}) encourages  our learned models to be robust against downgrade in audio-visual quality. Accuracy is consistently high for both video and audio downgrades, with no observable trend that severely compressed videos are incorrectly flagged as tampered. The tampering detection accuracy on the audio stream remains high except for the TNA dataset which starts dropping at $q = 8$. The fluctuation indicates that the visual stream compression factor is not correlated to accuracy.

Finally, Table.~\ref{tab:all} summarises the overall performance of our tamper detection system in term of Precision (positive predictive value), Recall (sensitivity) and F1 score. Similar to our observation in Fig.~\ref{fig:tamlen} the ASSAVID dataset with dated MPEG-2 codec and low video quality results in large tampering threshold $\epsilon_t$ (eqn.~\ref{eq:thres}) thus less sensitive to tampering. The more modern MPEG-4 encoded OLYMPICS dataset has a better balance between precision and recall with excellent performance also for the TNA dataset.

\begin{table}
    \centering
    \begin{tabular}{|c|c|c|c|}
    \hline
    Datasets  & Precision & Recall & F1 \\ \hline
    ASSAVID~\cite{messer2002automatic}   & 0.981      & 0.756 & 0.854 \\ \hline
    OLYMPICS  & 0.944      & 0.823 & 0.879 \\ \hline
    TNA       & 0.919      & 0.925 & 0.922 \\ \hline
    \end{tabular}
    \caption{Overall tamper detection performance of the 3 datasets in term of Precision, Recall and F1 score. Here, a true positive denotes a tampered video being correctly detected as tampered.}
    \label{tab:all}
    \squeezeup
\end{table}

%memory requirement, training time

\subsection{Blockchain Characteristics}

The smart contracts implementing APIs for the fetch (\ie search for and read a record) and write (commit new record) are deployed on an Ethereum PoA network with {\em geth} used for client API access and for block sealing at each of the three participating archives.  A further four block sealing nodes were present on the network during this trial deployment for debug and control purposes.  For purposes of evaluation the network was run privately between the nodes and genesis block configured with a 15 second default block sealing rate. The sealing rate limits the write (not fetch) speed of the network, so that worst-case transaction processing could be up to 15 seconds. The smart contract implemented fetch functionality via the {\em view} functions provided by the Solidity programming language, which only read and do not mutate the state, hence not requiring a transaction to be processed.

\begin{figure}
    \centering
    \includegraphics[width=\linewidth,height=5.5cm]{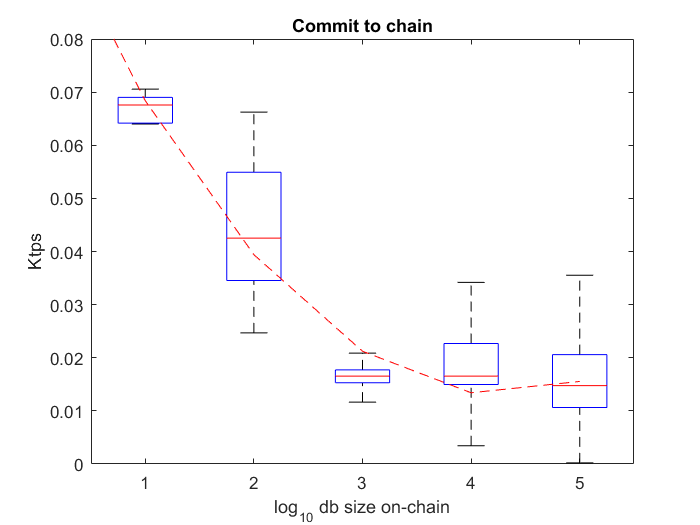}
    \includegraphics[width=\linewidth,height=5.5cm]{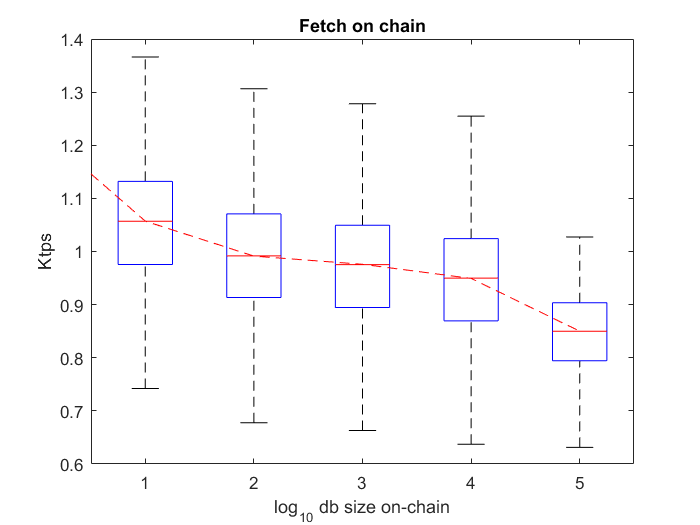}
    \caption{Performance scalability of smart contract transactions on our proof-of-authority Ethereum implementation.  Performance for fetching (top) and commiting (bottom)  video integrity records to the on-chain storage.  Measured for db sizes from $10^1-10^5$ reported as Ktps ($\times 10^3$ transactions per second). }
    \label{fig:dltperf}
        \squeezeupSmall
\end{figure}

To test the smart contract we devised a test of writing a number of records to the store, measuring the time taken to create and submit a transaction, and then reading a record a number of times to measure the time taken to read from the store. When writing the records we submitted them in batches of 1000 to ensure that the transactions would be accepted by the network. The transaction creation/submission performance is presented in Fig.~\ref{fig:dltperf} (top) and do not include the (up to) 15 second block sealing overhead since this is a constant dependent on the time at which the commit transaction is posted. The transactions were submitted in powers of 10, and then the read performance was measured. The read performance is presented in Fig.~\ref{fig:dltperf} (bottom).  %A dashboard snapshot of the Ethereum network is included in Fig.~\ref{fig:dash}.

%\begin{figure}
%    \centering
%    \includegraphics[width=\linewidth]{figs/live_network.png}
%    \includegraphics[width=\linewidth]{figs/test_network.png}
%    \caption{Ethereum dashboard snapshot of the live system during deployment.}
%    \label{fig:dash}
%\end{figure}

\section{Conclusion}

We have presented ARCHANGEL; a PoA blockchain-based service for ensuring the integrity of digital video across multiple archives.  Our key novelty of our approach is the fusion of computer vision and blockchain to immutably store temporal content hashes (TCHs) of video, rather than bit hashes such as SHA-256.  The TCHs are invariant to format shifting (changes in the video codec used to compress the content) but sensitive to tampering either due to frame dropout and truncation ({\em temporal} tamper) or frame corruption ({\em spatial} tamper).  This level of protection guards against accidental corruption due to bulk transcoding errors as well as direct attack. We evaluated an Ethereum deployment of our system across the national government archives of three sovereign nations.  We show that our system can detect spatial or temporal tampers of a few seconds within videos tens of minutes or even hours in duration.   

Currently ARCHANGEL is limited by Ethereum (rather than via design) to block sizes of 32Kb.  Our TCHs for a given clip are variable length bit sequences requiring approximately 1024 bits per minute to encode audio and video on-chain.  This effectively limiting the size of digital videos we can handle to 
 ~256 minutes (about 4 hours) maximum duration.   ARCHANGEL also has a relatively high overhead in off-chain storage in that a model of 100-200Mb must be stored per video clip; however this model size is frequently dwarfed by the video itself which may be gigabytes in length and presented minimal overhead in the context of the petabyte capacities of archival data stores.  One interesting question is how to archive such models; currently these are simply Tensorflow model snapshots but no open standards exist for serializing DNN models for archival purposes.  As deep learning technologies mature, and becomes further ingrained within autonomous decision making \eg by governments, it will become increasingly important for the community to devise such open standards to archive such models.
 Looking beyond ARCHANGEL's context of national archives, the fusion of content-aware hashing and blockchain holds further applications, \eg safeguarding journalistic integrity or to mitigate against fake videos (`deep fakes') via similar demonstration of content provenance.

\section*{Acknowledgement}

ARCHANGEL is a two year project funded by the UK Engineering and Physical Sciences Research Council (EPSRC). Grant reference EP/P03151X/1.

{\small
\bibliographystyle{ieee}
\bibliography{archangelws}
}

\end{document}